\documentclass[letterpaper, 10 pt, conference]{ieeeconf}  

\IEEEoverridecommandlockouts                              

\usepackage{graphics}
\usepackage{graphicx}
\usepackage{pdfpages}
\usepackage{color}
\usepackage{amsmath,amssymb}
\usepackage{mathtools}
\usepackage{color, soul}
\usepackage{multirow}
\usepackage{caption}
\usepackage{multicol}
\usepackage{xcolor}

\usepackage[font=small,labelfont=bf,margin=\parindent,tableposition=top]{caption}
\usepackage{subcaption}
\usepackage{titlesec}
\usepackage[colorlinks=true, allcolors=black]{hyperref}

\titlespacing*{\section}{0pt}{0.5\baselineskip}{0.2\baselineskip}
\titlespacing*{\subsection}{0pt}{0.35\baselineskip}{0.35\baselineskip}

\usepackage{algorithm,algcompatible,amsmath}
\algnewcommand\INPUT{\item[\textbf{Input:}]}%
\algnewcommand\OUTPUT{\item[\textbf{Output:}]}%
\algnewcommand\algorithmicreturn{\textbf{return}}
\algnewcommand\RETURN{\State \algorithmicreturn}

\usepackage{widetext}
\usepackage{cite}
\usepackage[font=small,labelfont=bf,margin=\parindent,tableposition=top]{caption}
\makeatletter
\let\NAT@parse\undefined
\makeatother

\title{\LARGE \bf
OptiRoute: A Heuristic-assisted Deep Reinforcement Learning Framework for UAV-UGV Collaborative Route Planning
}

\author{Md Safwan Mondal$^{1}$, Subramanian Ramasamy$^{1}$, Pranav Bhounsule$^{1}$
\thanks{*This work was supported by ARO grant
W911NF-14-S-003.}
\thanks{$^{1}$Md Safwan Mondal, Subramanian Ramasamy and 
Pranav A. Bhounsule are with the Department of Mechanical
and Industrial Engineering, University of Illinois Chicago, IL,
60607 USA.
        {\tt\small mmonda4@uic.edu, sramas21@uic.edu, pranav@uic.edu}}%
\thanks{
        {}}%
}

\begin{document}

\maketitle
\thispagestyle{plain}
\pagestyle{plain}

\begin{abstract}

Unmanned aerial vehicles (UAVs) are capable of surveying expansive areas, but their operational range is constrained by limited battery capacity. The deployment of mobile recharging stations using unmanned ground vehicles (UGVs) significantly extends the endurance and effectiveness of UAVs. However, optimizing the routes of both UAVs and UGVs, known as the UAV-UGV cooperative routing problem, poses substantial challenges, particularly with respect to the selection of recharging locations. Here in this paper, we leverage reinforcement learning (RL) for the purpose of identifying optimal recharging locations while employing constraint programming to determine cooperative routes for the UAV and UGV. Our proposed framework is then benchmarked against a baseline solution that employs Genetic Algorithms (GA) to select rendezvous points. Our findings reveal that RL surpasses GA in terms of reducing overall mission time, minimizing UAV-UGV idle time, and mitigating energy consumption for both the UAV and UGV. These results underscore the efficacy of incorporating heuristics to assist RL, a method we refer to as heuristics-assisted RL, in generating high-quality solutions for intricate routing problems. 

\end{abstract}

\section{Introduction}

Unmanned aerial vehicles (UAVs) have rapidly evolved as an emerging technology, finding profound applications in both military and civilian sectors \cite{liu2019cooperative, stolfi2021uav,tokekar2016sensor,wu2020cooperative}. They are critical for real-time sensing in scenarios like traffic monitoring \cite{puri2007statistical}, border security \cite{ozkan2021uav}, disaster management \cite{erdelj2017help} and forest fire surveillance \cite{yuan2015survey}, all of which demand continuous data transmission. However, a major limitation of UAVs in such persistent applications is their restricted operational time due to limited battery capacity. This challenge can be mitigated by leveraging the synergies of multi-agent systems that combine UAVs with unmanned ground vehicles (UGVs). This collaborative approach can enhance the overall task efficacy and prolong the UAVs' operational longevity \cite{maini2015cooperation,zhang2022cooperative}.

\begin{figure}[htbp]
\centering
\includegraphics[scale=0.42
]{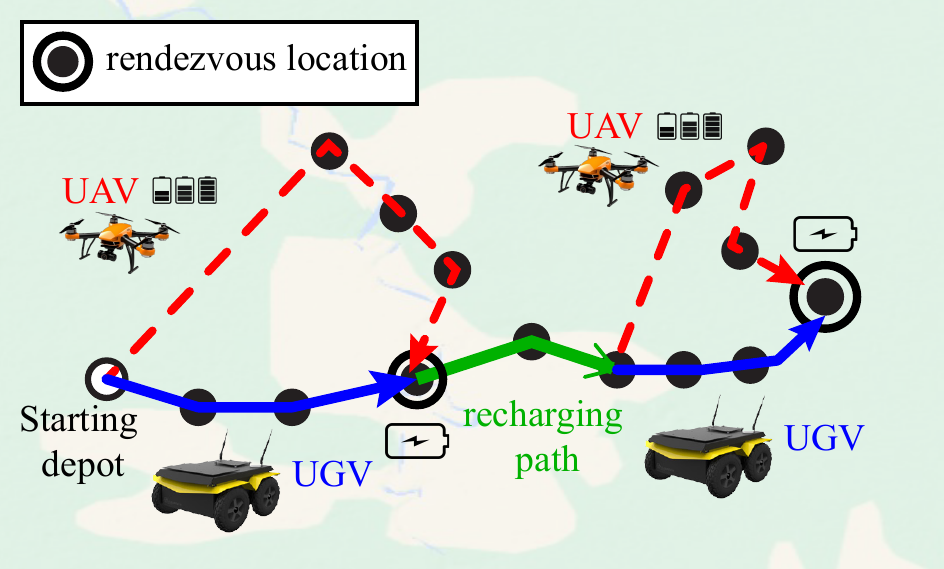}
\caption{Illustration of a fuel constrained UAV-UGV cooperative routing problem studied in this paper. DRL policy specifies rendezvous points, and heuristic planners outline optimal UAV-UGV paths to cover all task points.}
\label{problem}
\end{figure}

In this work, we address a \textit{cooperative routing problem} involving a heterogeneous team of a UAV and a UGV to visit a set of designated task nodes in the quickest possible time (see Fig. \ref{problem}). The UAV operates under limited battery life constraint and is supported by the UGV that acts as a mobile recharging depot. The UGV is also restricted in terms of speed and can only travel on the road network. The recharging process of UAV by UGV is not instantaneous which adds an additional layer of complexity to the problem. The challenge, therefore, is to meticulously strategize the routes for both the entities, ensuring that the UAV can effectively get recharged from the UGV to execute the mission optimally. This necessitates a comprehensive cooperative routing framework to optimally plan UAV and UGV routes while synchronizing their rendezvous recharging instances.

\subsection{Related works}

Extensive research has been carried out on different variants of UAV-UGV cooperative routing problem \cite{li2016hybrid, liu2020two, manyam2019cooperative, ramasamy2022coordinated}. Traditional methodologies, such as Mixed Integer Linear Programming (MILP) \cite{sundar2016formulations}, graph matching \cite{ghassemi2022multi}, and multi-echelon formulation \cite{liu2020two}, have previously been employed to tackle this type of combinatorial optimization problem (CO). In most studies, the cooperative routing is modelled as a variant of vehicle routing problem what makes it a NP-hard problem. These methods often fail to scale well with the number of the task points and lacks practical applicability due to the intricate combinatorial nature of the problem. Hence, in recent years, learning based methodologies have gained popularity as a promising solution to solve the combinatorial optimization problems.   

Paul et al. \cite{paul2022scalable,paul2023efficient} utilized an encoder-decoder based graph neural network to model mutlti-robot task allocation problem (MRTA) and showed learning based algorithms can produce quality solutions in significantly less computational time compare to the non-learning based baselines. Li et al. \cite{li2021deep} employed a deep reinforcement learning (DRL) method that leveraged attention mechanisms to tackle the heterogeneous capacitated vehicle routing problem. Their methodology outperformed traditional heuristic solutions in both quality and computational efficiency. Wu et al. \cite{wu2021reinforcement} investigated truck and drone based last mile delivery problem using a reinforcement learning (RL) paradigm. They split the optimization problem into customer clustering problem and routing components and applied an encoder-decoder framework with RL to resolve it. Fan et al. \cite{fan2022deep} employed a multi-head attention mechanism coupled with a DRL policy to design routes for an energy-limited UAV, however they assumed a fixed set of recharging locations for the UAV. 

In our prior research \cite{ramasamy2023solving, mondal2023bilevel}, we introduced a two-tiered optimization structure that streamlined the collaborative routing challenge. This framework determined the UGV route in the primary layer and the UAV route in the secondary layer using heuristic methods. We further demonstrated that optimizing the recharging instances between the UAV and UGV can enhance the cooperative routing solution's quality \cite{ramasamy2022heterogenous}. In this paper, we have coupled heuristic strategy-based route planners with a learning framework that has produced an optimal strategy for UAV-UGV recharging rendezvous planning to achieve the best cooperative route. Given the complexity of the task, we have harnessed the strengths of both heuristic and learning methods to navigate the scenario and have compared our solution with a non-learning baseline, which has shown substantial improvement in solution quality. To this end, we present the following novel contributions:\\
1. We have formulated the fuel constrained UAV-UGV cooperative routing problem as a Markov Decision Process (MDP) so that the recharging rendezvous  problem can be learned using a policy gradient RL approach. To the best of our knowledge, this is one of the first work to address fuel constrained UAV-UGV cooperative routing problem with RL.   \\
2. We have adopted an encoder-decoder based transformer architecture within the learning policy network that can help in more efficient data extraction from the scenario to select the appropriate refuel stop locations.\\
3. We have integrated the advantages of heuristics with RL framework by modelling the energy-constrained vehicle routing problem and solving it through constraint programming heuristics to get the UAV-UGV routes based on the recharging stop locations chosen by the RL policy. \\
4. The comparison with a non-learning baseline method shows the effectiveness and reliability of the proposed algorithm.  
\vspace{-3mm}
\section{Problem Formulation}

\subsection{Problem overview}

Given a set of task points $\mathcal{M} = \{m_0,m_1,...,m_n\}$ situated on a road network, the objective is to strategize a collaborative operation between a UAV $ A $ and a UGV $ G $ to optimally visit all the task points in the scenario. The vehicles have heterogeneous characteristics as the UAV boasts higher speed $v^a$ but operates within a restricted battery capacity $F^a$. It can, however, recharge its battery from the UGV, which travels at a slower pace $v^g$ exclusively on the road network. For recharging, the UAV docks onto the UGV's landing platform and once rejuvenated, it resumes its flight by taking off from UGV. The duration of this recharging is contingent upon the UAV's battery level prior to docking. The UAV's energy consumption profile, $ \mathcal{P}^a = 0.0461{v^a}^3-0.5834{v^a}^2-1.8761v^a+229.6 $\ being a inversely proportional of its velocity $v^a$ adds another layer of complexity to the problem as the hovering consumes more energy than transiting, hence discouraged during the task. Considering these multifaceted challenges, which largely revolve around the recharging interplay between the UAV and UGV, our objective is to develop an efficient recharging strategy ( \textbf{\textit{RendezvousPlanner}} ) which determines \textbf{where} and \textbf{when} the UAV can rendezvous with UGV for recharging. Based on this rendezvous, routes for both vehicles $\tau^a \in A, \tau^g \in G $ are constructed by the \textbf{\textit{UAVPlanner}} and \textbf{\textit{UGVPlanner}} to ensure comprehensive coverage of all task points in the shortest time span, while conserving energy and minimizing idle periods ( UAV hovering, UGV waiting ) during recharging sessions. 

\subsection{MDP formulation}

\begin{figure}[b]
\vspace{-2mm}
\centering
\includegraphics[scale=0.45
]{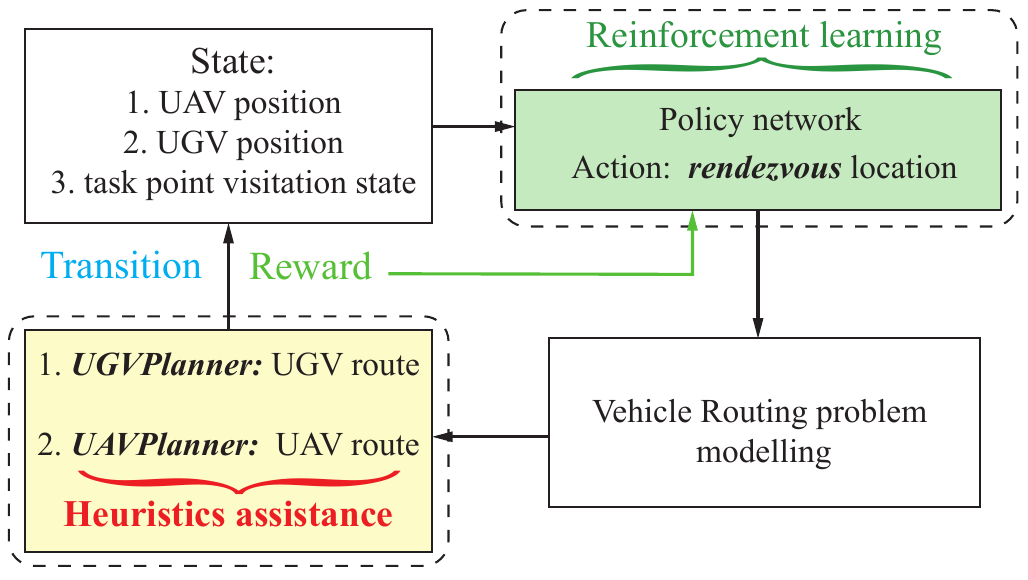}
\caption{MDP for heuristics-assisted UAV-UGV cooperative routing}
\label{MDP}
\vspace{1mm}
\end{figure}

The rendezvous planning  of this cooperative routing can be formulated as a MDP, since the refueling stop locations are chosen sequentially and its selection influences the state of entire task scenario (see Fig. \ref{MDP}) The components of the MDP can be defined as a tuple $ < \mathcal \mathcal{S}, \mathcal{A}, \mathcal{R}, \mathcal{T} > $, as following: \\ 
1) \textbf{State Space\ $(\mathcal{S})$:} In our MDP at any decision making step, the state of the environment $s_t \in \mathcal{S}$ is defined as, $ s_t = (\{x^a_t, y^a_t\} \in A, \{x^g_t, y^g_t\} \in G, \{x^i_t,y^i_t,d^i_t\}) \in \mathcal{M}) $, where, $\{x^a_t, y^a_t\}$ and $\{x^g_t, y^g_t\}$ represent the current coordinates of the UAV and UGV respectively and $\{x^i_t,y^i_t,d^i_t\}$ highlights the coordinates of task points and their visitation status, $d^i_t$ will be 1 if a task point is already visited or 0 if not. \\
2) \textbf{Action Space\ $(\mathcal{A})$:}  Selection of a task node as the refuel stop $m^r$ is considered to be the action $a_t \in \mathcal{A}$. Specifically, $a_t$ is defined as the index of the task node that is selected as rendezvous location at that decision making step, i.e, $a_t = i,  \ \forall \ m^r_i = {m_i} \in \mathcal{M}$.\\
3) \textbf{Reward\ ($\mathcal{R}$):} By keeping congruity with the objective of the cooperative routing problem to reduce total task completion time, we set the reward $r_t \in \mathcal{R}$ as the negative value of UAV route time $t^a_{route}$. This UAV route time is obtained by solving a vehicle routing problem (VRP) problem based on the rendezvous location $m^r_t$ chosen in the action $a_t$. The recharging time $t^a_{r}$ is calculated based on UAV's fuel consumption and subtracted from the reward function. Our reward function also deducts $t^a_{idle}$ and $t^g_{idle}$ to discourage UAV's hovering and UGV's waiting time period during rendezvous.
For effective reward shaping, we provide a significant positive bonus, 
$D$, upon task completion and impose a substantial penalty, 
$P$, for task failure. So, the reward $r^t$ can be written be as:  $r_t = r(s_t, a_t) = -t^a_{route} - t^a_r - t^a_{idle} - t^g_{idle} + D - P $.
Here, when the task is successfully completed, 
$D= 10000, P = 0$. In the case of task failure, 
$D = 0, P = 1000$. This reward structure promotes successful task completion and aids in achieving faster learning convergence.\\
4) \textbf{Transition $(\mathcal{T})$:} The next state $s_{t+1}$ in the transition function $\mathcal{T}(s_{t+1} | s_t, a_t )$ depends on the UAV route $\tau^a_t$ and UGV route $\tau^g_t$ obtained from route planners \textbf{\textit{UAVPlanner}} and \textbf{\textit{UGVPlanner}} based on selected action $a_t$. This is where we integrate routing heuristics (see section \ref{RH}) with RL framework for solving the cooperative route. In transition, the position of UAV and UGV is updated and the visitation state of the  mission points is updated based on the UAV and UGV routes as follow:
$ s_{t+1} = (\{x^a_{t+1}, y^a_{t+1}\} \in A, \{x^g_{t+1}, y^g_{t+1}\} \in G, \{x^i_{t+1},y^i_{t+1},d^i_{t+1}\}) \in \mathcal{M}) $, here
$d^i_{t+1} = 1, \ \text{ for unvisited mission points}\ m_i \in \tau^a_t | \tau^g_t$ and other elements is 0. Since stochasticity is not assumed the transition is considered to be deterministic.

\begin{figure*}[t]
\centering
\includegraphics[scale=0.43
]{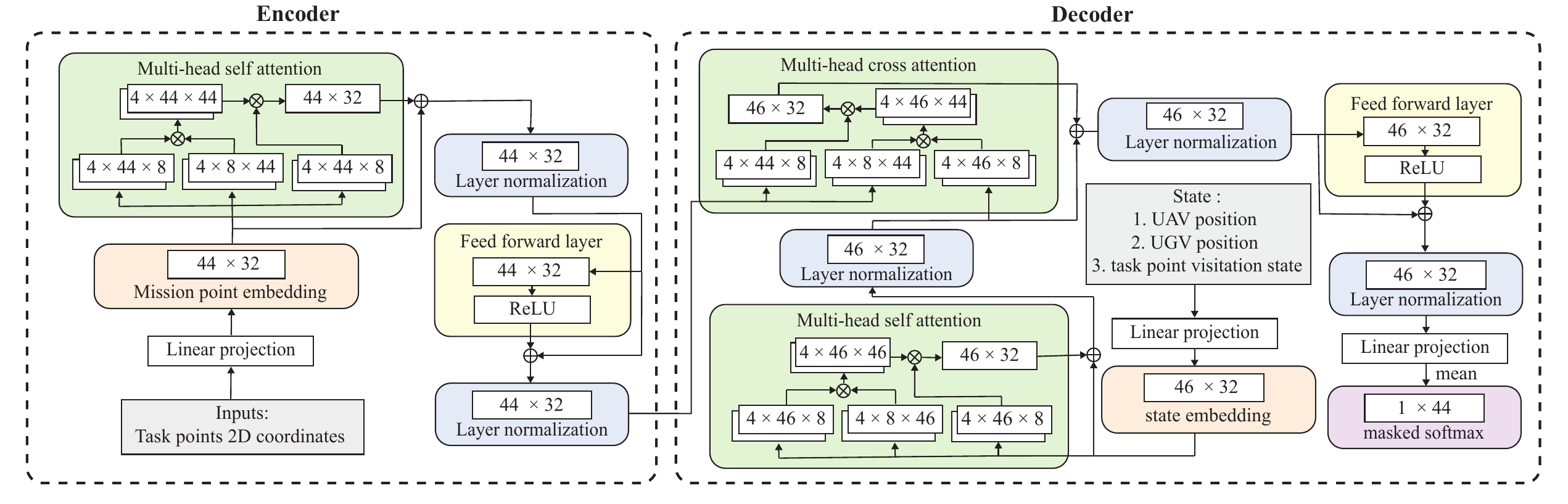}
\caption{Architecture of the proposed policy network. Encoder has one self attention layer to generate input scenario embedding, while decoder contains one self attention to yield \textbf{context} and one cross attention layer to generate attention score.}
\label{architecture}
\vspace{1mm}
\end{figure*}

\vspace{-1mm}
\section{Reinforcement learning framework}

In our study, we have implemented encoder-decoder based transformer architecture with RL algorithm to parameterize and learn the optimal policy $\pi_{\theta}$ with a trainable parameter  $\theta$. Starting from the initial state $s_0$, the policy $\pi_{\theta} = Pr(a_t | s_t, \theta_t = \theta) $ takes action $a_t$ to select appropriate refuel stops based on the scenario state $s_t$ until the terminal state $s_\tau$ is reached. The final solution of the policy network consists of a series of refuel stops chosen sequentially which can be represented as joint probability distribution as follow: 
\begin{equation}
Pr(s_\tau | s_0) = \prod\limits_{t=0}^{\tau-1} \pi_\theta(a_t | s_t) Pr(s_{t+1} | s_t, a_t)
\end{equation}

\subsection{Encoder-Decoder Transformer architecture}

Our policy network is consists of two main components: an encoder and a decoder. The encoder translates the task point coordinates into a more nuanced, high-dimensional embedding for better feature extraction. On the other hand, the decoder in the policy network $\pi_{\theta}$, discerns suitable rendezvous locations based on the encoder embedding and the contextual information extracted from the scenario's current state. A detailed description of the policy architecture (see Fig. \ref{architecture}) is explained here: 

\subsubsection{Encoder} We have utilized a multi head attention (MHA) mechanism \cite{bahdanau2014neural,vaswani2017attention} in the encoder for higher dimensional representation of the raw features of the problem instance. The encoder takes the  normalized 2D coordinates of the task points as an input $ X = (o_i = \{x_i, y_i\}, \forall \ m_i \in \mathcal{M} ) $ and linearly projects them to a higher dimension space with dimension $d_h = 32$ to create input embedding $h^0_i$.  This input embedding  is subsequently transformed using a single-layer multi-head self-attention mechanism. This transformation process yields  $h^L_i$, allowing a more detailed understanding of the relationships among the task points. The Attention layer calculates three vectors, \textit{query}, \textit{key} and \textit{value} from the input node embedding $h^0_i$ with the dimension of \textit{query}/\textit{key}, \textit{value} as, $d_q/d_k = d_v = \frac{d_h}{M} $, here $M = 4$, number of attention heads. For each head $j \in {1,2, ...M} $, the attention scores $Z_j$ are calculated between \textit{query} $q_{i,j}$ and \textit{key} $k_{i,j}$ and concatenated together to the original embedding dimension $h^0_i$. The calculations are shown here:
\begin{gather}
q_{i,j} = h^0_iW^q_j,\ k_{i, j} = h^0_iW^k_j,\ v_{i, j} = h^0_iW^v_j \\
Z_j = \text{softmax}\left( \frac{q_{i,j}{k_{i, j}}^{T}} {\sqrt{d_k}} \right)v_{i, j} \\
h^L_i = MHA( h^0_i) = \text{Concat}(Z_1, Z_2, ..., Z_j) 
\end{gather}

Here, $q_{i,j}, k_{i, j} \ \text{and} \ v_{i, j}$ are the \textit{query}, \textit{key} and \textit{value} respectively in head $j$ and $W^q_j \in \mathbb{R}^{d_h \times d_q}, W^k_j \in \mathbb{R}^{d_h \times d_k}$ and $W^v_j \in \mathbb{R}^{d_h \times d_v} $ are the trainable parameter matrices in the attention layer. The self attention output $h^L_i$ has a residual skip connection around them, followed by Layer-Normalization (LN). Then for the richer interpretation of feature space, the  output $h^r_i$ is refined through a Feed Forward (FF) layer followed by another residual skip connection and Layer-Normalization that yields the encoder embedding $h^f_i$ of the problem instance which is leveraged in the decoder.
\begin{gather}
h^r_i = LN(h^0_i + MHA(h^0_i)), \\
h^f_i = LN( h^r_i + FF(\text{ReLU}(h^r_i)))  
\end{gather}

\subsubsection{Decoder:}
During each decision-making step, the decoder determines the probability of selecting each available node as an action based on the encoder's node embedding $h^f_i$ and a \textbf{context} vector, which provides insights into the current scenario state. The decoder employs multi head self attention layer to extract state features as the context $H^{con}_t$ which is achieved by linear transformation of current state $s_t$ into $H^0_i$ and subsequent processing through multi head self attention layer followed by skip residual connection and Layer-Normalization as shown in following equations: 
\begin{gather}
H^0_i = \mbox{linear}(s_t), \ H^{con}_t = LN(H^0_i + MHA(H^0_i))
\end{gather}
Then this context vector $H^{con}_t$ is treated as the \textit{Query} and the encoder node embedding $h^f_i$ is considered as the \textit{Key}/\textit{Value} pair for a multi-head heterogeneous cross attention layer to generate $H^{attn}_t$.
\begin{gather}
H^{attn}_t = MHA( H^{con}_tW^Q_j,\ h^f_iW^K_j, \ h^f_iW^V_j  ) 
\end{gather}

Following the original transformation network \cite{vaswani2017attention, bahdanau2014neural}, the cross attention score $H^{attn}_t$ has a skip connection and Layer-Normalization. This is succeeded by a Feed Forward (FF) layer and another Layer-Normalization, further refining it to $H^{f}_t$. At the final step, the attention score is linearly projected to the action space dimension and is then averaged across the nodes.
\begin{gather}
H^r_t = LN( H^{attn}_t + H^{con}_t) \\
H^{f}_t = LN(H^r_t + FF(\text{ReLU}(H^r_t))) \\
H^{out}_t = \text{mean}(\text{linear}(H^{f}_t))
\end{gather}

An important step in this decoding step is to mask the invalid action nodes by setting its attention score to negative infinity. In our case, any task point that is out of reach of UAV and UGV from its current position and fuel capacity is considered as invalid at that action step. Finally, a softmax layer outputs the probability distribution over the action nodes, $Pr(a_t | s_t, \theta_t = \theta) =  \text{softmax}(H^{out}_t) $, indicating the likelihood of each node to be selected  as the refueling point. We have utilized sampling strategy to sample the node as action according to their probability in the decoding stage. 

\subsection{Training method} 
The training of our policy network leverages the REINFORCE policy gradient method, a seminal approach in the realm of reinforcement learning. At its core, REINFORCE seeks to optimize policies by directly adjusting policy parameters in the direction that improves expected rewards. The training process ( see Algorithm \ref{algorithm:REINFORCE} ) begins by initializing the policy parameters, often randomly. We execute an episode under the current policy, producing state, action, and reward sequences. Each step's return is calculated, and the policy parameters are adjusted based on the gradient of the log probability of actions, scaled by these returns. This iterative process makes actions with higher returns more likely in future episodes. The method's strength lies in its model-free nature, eliminating the need for explicit knowledge of the environment's dynamics. However, in our study future enhancements may include baseline networks with algorithms  like Proximal Policy Optimization (PPO) or Actor-Critic method for better convergence.

\begin{algorithm}[htbp]
    \small
    \caption{Policy network training using REINFORCE}
    \begin{algorithmic}[1]
        \INPUT  Policy network $\pi_\theta$, epochs $E$, batches $B$, episode length $T$, learning rate $\alpha$, discount factor $\gamma$
        \OUTPUT Trained policy network $\pi_{\theta^{'}}$
        \STATE Initialize policy $\pi_{\theta}$ parameters $\theta$
        \FOR{epoch in $1 \dots E$}
                \State Initialize trajectory set $\mathcal{E} = [ \ ]$
                \FOR{instance in $1 \dots B$}
                    \State Initialize $s_0$ and $t \leftarrow 0$
                    \State Initialize trajectory $\tau = [ \ ]$
                    \WHILE{$t < T$}
                        \State Get action $a_t \sim \pi_\theta(a_t | s_t)$
                        \State Obtain $r_t$ and $s_{t+1}$
                        \State Store $(s_t, a_t, r_t)$ in $\tau$, $ t = t + 1$
                    \ENDWHILE
                    
                    \State Append $\tau$ to $\mathcal{E}$
                \ENDFOR
                \State Compute gradient:
                \[
                \nabla_\theta J = \frac{1}{B} \sum_{\tau \in \mathcal{E}} \sum_{(s,a,r) \in \tau} \nabla_\theta \log \pi_\theta(a | s) \sum_{t'=t}^T \gamma^{t'-t} r_{t'}
                \]
                \State Update policy parameters: $\theta \leftarrow \theta + \alpha \nabla_\theta J$
        \ENDFOR
    \end{algorithmic}
    \label{algorithm:REINFORCE} 
\end{algorithm}

\vspace{-1.5mm}

\section{Routing Heuristics}
\label{RH}
Based on the refuel stops selected by the policy network (\textit{\textbf{RendezvousPlanner}}), we can model vehicle routing problem with constraints to obtain the UGV \& UAV routes  which in turn give us the reward of our MDP in the learning framework. Since the UGV is slower than the UAV, we followed the `UGV first, UAV second' heuristics approach of cooperative routing to construct the UGV route as the initial step and then construct the UAV route based on the spatio-temporal components of UGV route.

\subsubsection{UGV route planner} The \textbf{\textit{UGVPlanner}} generates the UGV path by connecting the UGV's current position to the chosen refuel stops. Since the UGV is confined to the road network, the waypoints between its present position and the refuel stop provide its route's spatial aspect. Given the UGV's consistent speed, its temporal journey can also be determined. To align with the UAV for refueling, the UGV may wait at the refuel station if necessary, however waiting at the refuel stops is discouraged to reduce idle period. The arrival time of the UGV at the refuel stop is fed to \textbf{\textit{UAVPlanner}} which models the UAV routing as the energy constrained vehicle routing problem with time windows (E-VRPTW). 

\subsubsection{UAV route planner} The E-VRPTW formulation can be depicted using graph theory. Here, the task points act as the vertices $V = \{S, 0, 1, ... D\}$ and $ E= \{(i, j) \, \| \ i, j \, \in \, V, i \neq j \}$ denotes the edges connecting vertices $i$ and $j$. We assign non-negative arc cost between vertices $i$ and $j$ as $t_{ij}$ (traversal time) and the decision variable as $x_{ij}$ that indicates whether a vehicle transits from $i$ to $j$. The UAV commences it's journey at starting point $S$, visits the task points and when needed it terminates its route to recharge from the UGV at refuel stop $D$, which is bounded by a time-window due to UGV's slower pace. The objective (Eq. \ref{eq:1}) is minimizing the cumulative travel duration while dropping least number of task points by penalizing each task point drop with a large penalty $P$; here $y_i$ is a binary variable that's set to 1 if a point is visited or 0 otherwise. We've established energy constraint in Eqs. \ref{eq:2} - \ref{eq:3} to make sure that the UAV never runs out of its fuel and its fuel consumption follows the UAV's power consumption profile during traversal (Eq. \ref{eq:4}). The time-window condition in Eq. \ref{eq:5} makes the UAV visit the UGV only after its arrival at refuel stop $D$. Eq. \ref{eq:6} says that the cumulative arrival time at $j^{th}$ node is equal to the sum of cumulative time at the node $i$, $t_i$ and the travel time between them $t_{ij}$. In both Eq. \ref{eq:4} \& Eq. \ref{eq:6} we have applied Miller-Tucker Zemlin (MTZ) formulation \cite{miller1960integer} by adding large constant $L_1, L_2$ for sub-tour elimination in the UAV route. The other generic constraints for the UAV route like flow conservation are detailed in our previous work \cite{mondal2023cooperative}. 
\vspace{2 mm}

\text{Objective: }
\begin{equation}
 \min \sum_i \sum_j t_{ij} x_{i j} + P \sum_i (1 - y_i) \quad \forall i, j \in V  \label{eq:1}
\end{equation}

\text{Energy constraints:} 
\begin{equation}
f^a_j= F^a, \quad j \in D \label{eq:2}
\end{equation}
\begin{equation}
0 \leq f^a_j \leq F^a, \quad \forall j \in V \setminus \{S,D\} \label{eq:3}
\end{equation}
\begin{align}
    f^a_j &\leq f^a_i-\left(\mathcal{P}^a(v^a)t_{i j}x_{i j}\right) \nonumber \\
    &+ L_1\left(1-x_{i j}\right), \quad \forall i \in V, j \in   V \setminus \{S,D\} \label{eq:4}
\end{align}

\text{Time window constraints:}
\begin{equation}
t_{j,start} \leq t_j \leq t_{j,end} \ , \quad \forall j \in D \label{eq:5}
\end{equation}
\begin{equation}
t_j \geq t_i+\left(t_{i j} x_{i j}\right)-L_2\left(1-x_{i j}\right), \quad \forall i \in V, j \in V \label{eq:6}
\end{equation}

The UAV route is calculated by the \textbf{\textit{UAVPlanner}} by solving the above E-VRPTW formulation with Google
OR-Tools \texttrademark \ CP-SAT solver \cite{ORtools} that uses constrained programming (CP) and we leverage \texttt{TABU SEARCH} heuristics in the solver for avoiding the local optimal solution.

Distinct from earlier cooperative routing studies \cite{maini2015cooperation, liu2020two}, our approach leverages mobile recharging, turning recharging duration into opportunities for mission point visitation. Once the UAV-UGV route is determined, the UAV's refueling time is computed based on its in-flight fuel consumption. This refueling time is projected  as a distance along road network where UGV transports the UAV, effectively recharging it while concurrently visiting task points along the road. The end location of the recharging process will be the next take off point of the UAV and the route time, recharging time, UAV hovering time and UGV waiting time are fed back to the MDP reward function and we will move forward to the next cycle in MDP.

\vspace{-1.5mm}
\section{Simulation Results}
\begin{figure*}[t]
\centering
\includegraphics[scale=0.41]{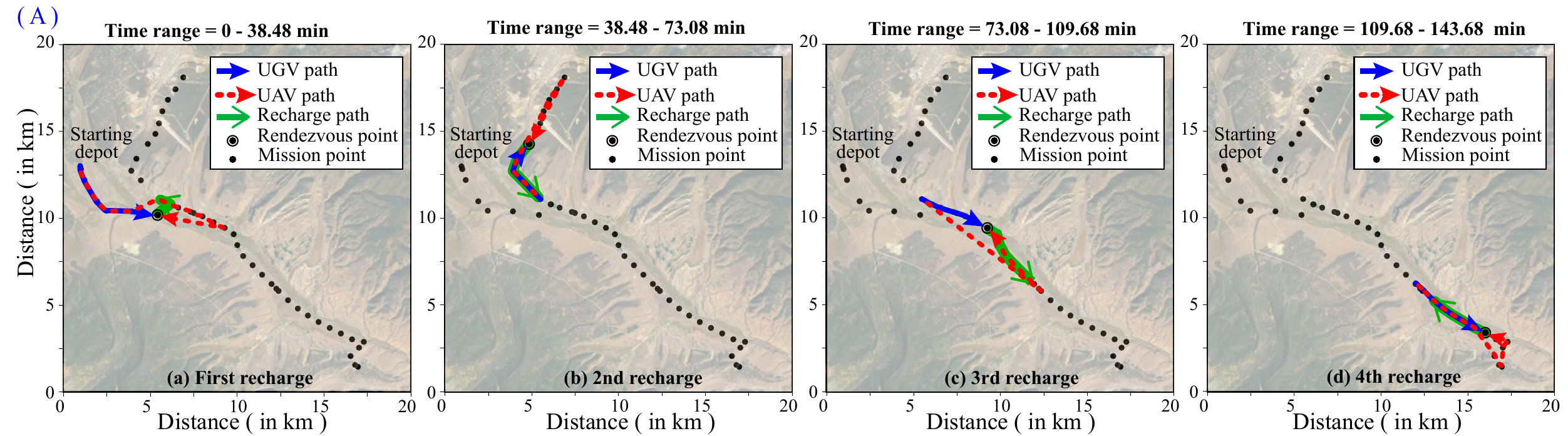} 
\includegraphics[scale=0.41
]{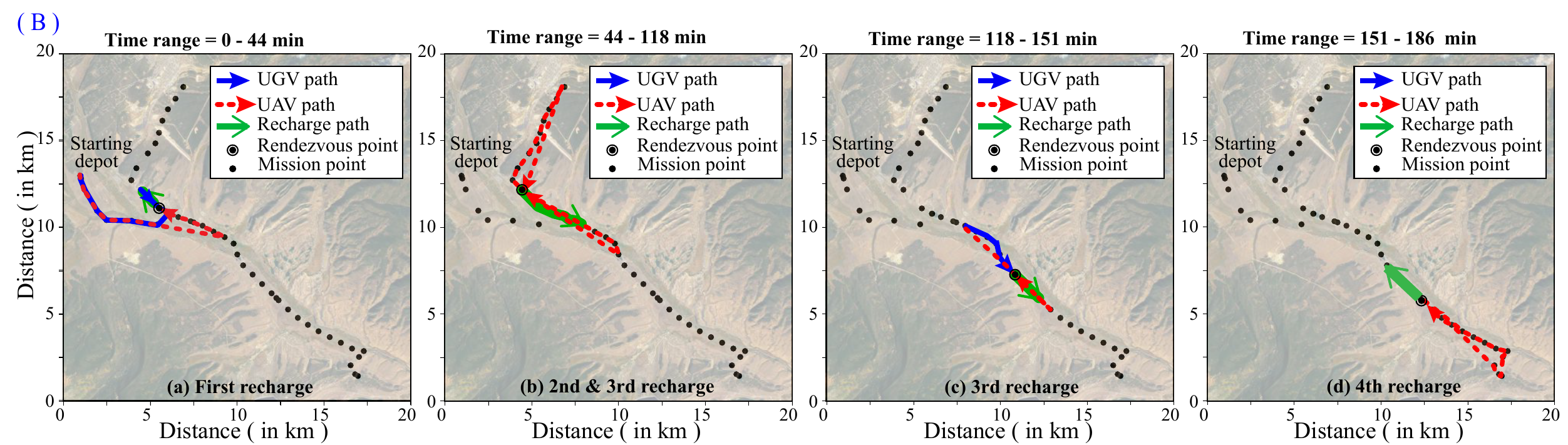}
\caption{Cooperative routes of UAV-UGV at recharging instances A) DRL method B) GA method. The route animation can be found at \url{http://tiny.cc/x55bvz}}
\label{fig:routes}
\end{figure*}

In this section, we simulate the fuel-constrained cooperative routing of a UAV-UGV team across a 20 km $ \times$ 20 km real world site. Starting from a depot, the team aims to visit task points shown as black dots in Fig. \ref{fig:routes}. The UAV and UGV moves at $v^a = 10$ m/s and $v^g = 4.5$ m/s speed respectively and UAV has a fuel capacity of 287.7 kJ. The recharging instances between UAV and UGV are determined by \textit{\textbf{RendezvousPlanner}} and their route is determined by \textbf{\textit{UAVPlanner}} and \textit{\textbf{UGVPlanner}} respectively as previously discussed. The goal is to accomplish the task in minimum possible time while ensuring that the UAV and UGV spend minimal idle time hovering or waiting during their recharging rendezvous instances. We evaluate the following metrics: 1) the total task completion time as the main metric 2) the idle time spent percentage (\text{idle time} \big/ \text{task completion time}) by UAV and UGV during rendezvous 3) the UAV and UGV energy consumption metrics during the entire task period. Given the unique complexity and specificity of our challenge, there are no standard benchmarks available, nor could an exact solution be discerned. Hence we compare our RL based results with results derived from a Genetic algorithm (GA) based non-learning baseline. The details about the implementation of GA for this cooperative routing can be found in our previous study \cite{ramasamy2022heterogenous}.  All computational procedures were executed using a Python simulation on a 3.7 GHz Intel Core i9 processor, equipped with 32 GB RAM, operating on a 64-bit system.

\begin{figure}[htb]
\centering
\includegraphics[scale=0.45]{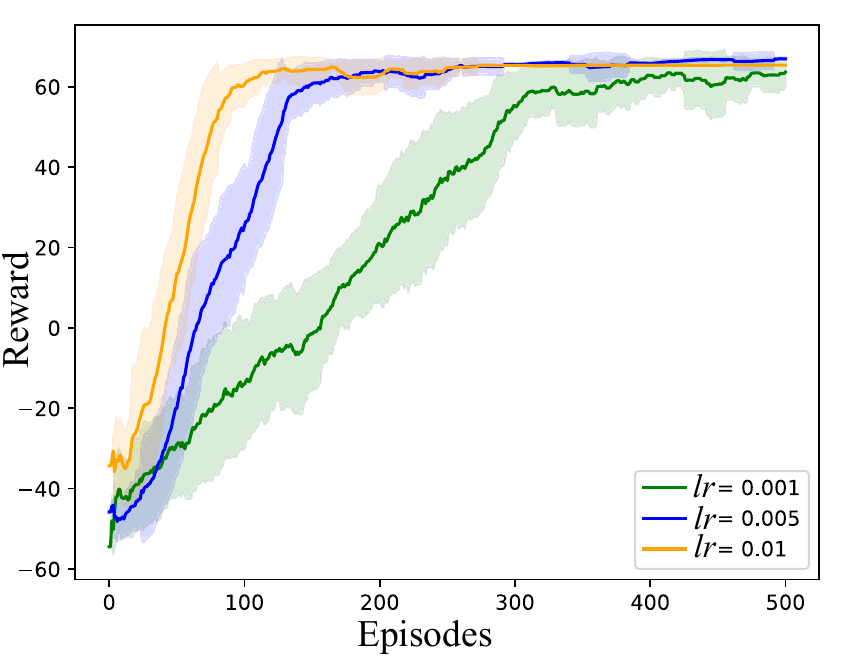}
\caption{Mean episode reward in training process }
\label{fig:reward}
\end{figure}

The training of RL is conducted in 500 episodes with different learning rates 
$lr = 0.001, 0.005, 0.01$. The learning rate significantly influences the RL algorithm's convergence speed. As illustrated in Fig. \ref{fig:reward}, a lower $lr = 0.001$ results in a prolonged convergence, whereas an aggressive  $lr = 0.01$ converges swiftly but tends to overfit, compromising policy exploration. The intermediate $lr = 0.005$ strikes a balance between exploration and exploitation, emerging as the most effective for deriving an optimal policy. Consequently, we adopt the $\pi_\theta$ policy trained with $lr = 0.005$ for coordinating the UAV-UGV rendezvous in our task scenario.

\vspace{-2mm}
\vspace{2.5mm}

\begin{table}[htbp]
\vspace{5mm}
\renewcommand{\arraystretch}{0.1}
\centering
\caption{Comparison of Routing metrics }
\begin{tabular}{p{4cm}  p{1.6cm}  p{1.6cm}} 
 \hline \\[1ex]
 \textbf{Route Metrics} & \textbf{DRL method}  & \textbf{GA method} \\ [1ex] 
\hline\\[1ex]
 Total task completion time (min) & 143.00  & 186.00  \\ [1.5ex]
 Mission points dropped & 0 & 0\\[1.5ex]
 UAV travel time (min) & 84.72  & 99  \\ [1.5ex]
 UAV hovering time (min) & 2.41 & 4 \\ [1.5ex]
 UAV hovering energy cost (kJ) & 33.26 & 55.20 \\ [1.5ex]
UAV total energy cost (kJ)  & 1039.73 & 1231.32 \\ [1.5ex]
UGV travel time (min) & 71.92  & 52.00  \\ [1.5ex]
 UGV waiting time (min) & 15.22 & 61.00 \\ [1.5ex]
 UGV waiting energy cost (kJ) & 325.30 & 1304.06 \\ [1.5ex]
UGV total energy cost (kJ)  & 19347.13 & 19911.56 \\ [1ex]
\hline\\
\end{tabular}
\label{route metrics}
\vspace{-8pt}
\end{table}

Although, both the GA and DRL frameworks are able to come up with routing strategy to cover entire task scenario without drooping any task nodes, Fig. \ref{fig:routes} shows the difference of the cooperative routes obtained by them. DRL optimally determines the recharging rendezvous between the UAV and UGV, ensuring even-handed utilization of both in the routing process. In contrast to the GA approach, DRL's chosen refuel stops are further from previous rendezvous points. Consequently, the UGV travels longer to rendezvous locations hence doesn't endure prolonged wait times at refuel stops for UAV to land. From table \ref{rendezvous metrics} reveals that with strategic rendezvous planning, DRL requires only 4 recharging instances to complete the task, while GA demands 5. More recharging events add extra recharging and UAV detour times, leading to a prolonged overall task completion time (see table \ref{route metrics}). Also, owing to the less detours and less hovering UAV total energy consumption is 15\% lesser in DRL method. 
\vspace{4mm}

\begin{table}[htbp]
\renewcommand{\arraystretch}{0.1}
\centering
\caption{Comparison of Rendezvous metrics}
\begin{tabular}{p{4cm}  p{1.6cm}  p{1.6cm}} 
 \hline \\[1ex]

 \textbf{Rendezvous Metrics} & \textbf{DRL method}  & \textbf{GA method} \\ [1ex] 
\hline\\[1ex]
No. of recharge stops & 4 & 5 \\[1.5ex]
Recharge time (min) & 55.87  & 73.00  \\ [1.5ex]
UAV idle time (\%) & 1.68 & 2.15 \\ [1.5ex]
UGV idle time (\%) & 10.64 & 32.79 \\ [1ex]
\hline\\
\end{tabular}
\label{rendezvous metrics}
\vspace{-12pt}
\end{table}

The impact of the rendezvous policy can be clearly understood in table \ref{rendezvous metrics}. In DRL rendezvous policy, UAV and UGV spend less amount of time in remain idle either in hovering or waiting. Particularly, the UGV idle time is 3 times less in the DRL method what fulfills the objective of this work. Hence we can say, DRL method is able to learn the given environment to figure out the optimal recharging policy between the UAV and UGV which ultimately impacts the overall the cooperative route positively by reducing the task completion time and vehicle idleness and energy consumption.    
\vspace{-4pt}
\section{Conclusion \& Future work}
In this study, we introduced a novel framework that combined heuristic guidance with deep reinforcement learning to address the fuel-constrained UAV-UGV cooperative routing problem. We demonstrated the effectiveness of using an encoder-decoder transformer architecture for making rendezvous decisions, while a heuristic planner efficiently devised routes for the UAV and UGV to minimize total mission time. By harnessing the decision-making capabilities of reinforcement learning and the route generation through heuristics, our approach aptly navigated the complexities of such a problem. With our DRL-based framework, we achieved a 23\% reduction in task completion time compared to the non-learning GA baseline. Additionally, through strategic rendezvous planning, we significantly minimized the UAV-UGV idle periods, leading to a considerable decrease in their energy consumption. In future, we plan to incorporate batch processing in RL to enhance training, aiming for a more robust solution. This will allow us to make more broader claim about the viability of our approach. Furthermore, we intend to adapt our framework for dynamic routing to account for unpredictable variations.


\clearpage
\addtolength{\textheight}{-12cm}

\addtolength{\textheight}{12cm}   





\bibliographystyle{unsrt}   
\bibliography{ICRA24}

\begin{thebibliography}{10}

\bibitem{liu2019cooperative}
Yao Liu, Zhihao Luo, Zhong Liu, Jianmai Shi, and Guangquan Cheng.
\newblock Cooperative routing problem for ground vehicle and unmanned aerial
  vehicle: The application on intelligence, surveillance, and reconnaissance
  missions.
\newblock {\em IEEE Access}, 7:63504--63518, 2019.

\bibitem{stolfi2021uav}
Daniel~H Stolfi, Matthias~R Brust, Gr{\'e}goire Danoy, and Pascal Bouvry.
\newblock Uav-ugv-umv multi-swarms for cooperative surveillance.
\newblock {\em Frontiers in Robotics and AI}, 8:616950, 2021.

\bibitem{tokekar2016sensor}
Pratap Tokekar, Joshua Vander~Hook, David Mulla, and Volkan Isler.
\newblock Sensor planning for a symbiotic uav and ugv system for precision
  agriculture.
\newblock {\em IEEE transactions on robotics}, 32(6):1498--1511, 2016.

\bibitem{wu2020cooperative}
Yu~Wu, Shaobo Wu, and Xinting Hu.
\newblock Cooperative path planning of uavs \& ugvs for a persistent
  surveillance task in urban environments.
\newblock {\em IEEE Internet of Things Journal}, 8(6):4906--4919, 2020.

\bibitem{puri2007statistical}
Anuj Puri, KP~Valavanis, and M~Kontitsis.
\newblock Statistical profile generation for traffic monitoring using real-time
  uav based video data.
\newblock In {\em 2007 Mediterranean Conference on Control \& Automation},
  pages 1--6. IEEE, 2007.

\bibitem{ozkan2021uav}
Omer Ozkan and Muhammed Kaya.
\newblock Uav routing with genetic algorithm based matheuristic for border
  security missions.
\newblock {\em An International Journal of Optimization and Control: Theories
  \& Applications (IJOCTA)}, 11(2):128--138, 2021.

\bibitem{erdelj2017help}
Milan Erdelj, Enrico Natalizio, Kaushik~R Chowdhury, and Ian~F Akyildiz.
\newblock Help from the sky: Leveraging uavs for disaster management.
\newblock {\em IEEE Pervasive Computing}, 16(1):24--32, 2017.

\bibitem{yuan2015survey}
Chi Yuan, Youmin Zhang, and Zhixiang Liu.
\newblock A survey on technologies for automatic forest fire monitoring,
  detection, and fighting using unmanned aerial vehicles and remote sensing
  techniques.
\newblock {\em Canadian journal of forest research}, 45(7):783--792, 2015.

\bibitem{maini2015cooperation}
Parikshit Maini and PB~Sujit.
\newblock On cooperation between a fuel constrained uav and a refueling ugv for
  large scale mapping applications.
\newblock In {\em 2015 international conference on unmanned aircraft systems
  (ICUAS)}, pages 1370--1377. IEEE, 2015.

\bibitem{zhang2022cooperative}
Mingjia Zhang, Huawei Liang, and PengFei Zhou.
\newblock Cooperative route planning for fuel-constrained ugv-uav exploration.
\newblock In {\em 2022 IEEE International Conference on Unmanned Systems
  (ICUS)}, pages 1047--1052. IEEE, 2022.

\bibitem{li2016hybrid}
Jianqiang Li, Genqiang Deng, Chengwen Luo, Qiuzhen Lin, Qiao Yan, and Zhong
  Ming.
\newblock A hybrid path planning method in unmanned air/ground vehicle
  (uav/ugv) cooperative systems.
\newblock {\em IEEE Transactions on Vehicular Technology}, 65(12):9585--9596,
  2016.

\bibitem{liu2020two}
Yao Liu, Zhong Liu, Jianmai Shi, Guohua Wu, and Witold Pedrycz.
\newblock Two-echelon routing problem for parcel delivery by cooperated truck
  and drone.
\newblock {\em IEEE Transactions on Systems, Man, and Cybernetics: Systems},
  51(12):7450--7465, 2020.

\bibitem{manyam2019cooperative}
Satyanarayana~G Manyam, Kaarthik Sundar, and David~W Casbeer.
\newblock Cooperative routing for an air--ground vehicle team—exact
  algorithm, transformation method, and heuristics.
\newblock {\em IEEE Transactions on Automation Science and Engineering},
  17(1):537--547, 2019.

\bibitem{ramasamy2022coordinated}
Subramanian Ramasamy, Jean-Paul~F Reddinger, James~M Dotterweich, Marshal~A
  Childers, and Pranav~A Bhounsule.
\newblock Coordinated route planning of multiple fuel-constrained unmanned
  aerial systems with recharging on an unmanned ground vehicle for mission
  coverage.
\newblock {\em Journal of Intelligent \& Robotic Systems}, 106(1):30, 2022.

\bibitem{sundar2016formulations}
Kaarthik Sundar, Saravanan Venkatachalam, and Sivakumar Rathinam.
\newblock Formulations and algorithms for the multiple depot, fuel-constrained,
  multiple vehicle routing problem.
\newblock In {\em 2016 American Control Conference (ACC)}, pages 6489--6494.
  IEEE, 2016.

\bibitem{ghassemi2022multi}
Payam Ghassemi and Souma Chowdhury.
\newblock Multi-robot task allocation in disaster response: Addressing dynamic
  tasks with deadlines and robots with range and payload constraints.
\newblock {\em Robotics and Autonomous Systems}, 147:103905, 2022.

\bibitem{paul2022scalable}
Steve Paul and Souma Chowdhury.
\newblock A scalable graph learning approach to capacitated vehicle routing
  problem using capsule networks and attention mechanism.
\newblock In {\em International Design Engineering Technical Conferences and
  Computers and Information in Engineering Conference}, volume 86236, page
  V03BT03A045. American Society of Mechanical Engineers, 2022.

\bibitem{paul2023efficient}
Steve Paul, Wenyuan Li, Brian Smyth, Yuzhou Chen, Yulia Gel, and Souma
  Chowdhury.
\newblock Efficient planning of multi-robot collective transport using graph
  reinforcement learning with higher order topological abstraction.
\newblock {\em arXiv preprint arXiv:2303.08933}, 2023.

\bibitem{li2021deep}
Jingwen Li, Yining Ma, Ruize Gao, Zhiguang Cao, Andrew Lim, Wen Song, and Jie
  Zhang.
\newblock Deep reinforcement learning for solving the heterogeneous capacitated
  vehicle routing problem.
\newblock {\em IEEE Transactions on Cybernetics}, 52(12):13572--13585, 2021.

\bibitem{wu2021reinforcement}
Guohua Wu, Mingfeng Fan, Jianmai Shi, and Yanghe Feng.
\newblock Reinforcement learning based truck-and-drone coordinated delivery.
\newblock {\em IEEE Transactions on Artificial Intelligence}, 2021.

\bibitem{fan2022deep}
Mingfeng Fan, Yaoxin Wu, Tianjun Liao, Zhiguang Cao, Hongliang Guo, Guillaume
  Sartoretti, and Guohua Wu.
\newblock Deep reinforcement learning for uav routing in the presence of
  multiple charging stations.
\newblock {\em IEEE Transactions on Vehicular Technology}, 2022.

\bibitem{ramasamy2023solving}
Subramanian Ramasamy, Md~Safwan Mondal, Jean-Paul~F Reddinger, James~M
  Dotterweich, James~D Humann, Marshal~A Childers, and Pranav~A Bhounsule.
\newblock Solving vehicle routing problem for unmanned heterogeneous vehicle
  systems using asynchronous multi-agent architecture (a-teams).
\newblock In {\em 2023 International Conference on Unmanned Aircraft Systems
  (ICUAS)}, pages 95--102. IEEE, 2023.

\bibitem{mondal2023bilevel}
Md~Safwan Mondal, Subramanian Ramasamy, and Pranav Bhounsule.
\newblock A bilevel optimization framework for fuel-constrained uav-ugv
  cooperative routing: Planning and experimental validation.
\newblock {\em arXiv preprint arXiv:2303.02315}, 2023.

\bibitem{ramasamy2022heterogenous}
Subramanian Ramasamy, Md~Safwan Mondal, Jean-Paul~F Reddinger, James~M
  Dotterweich, James~D Humann, Marshal~A Childers, and Pranav~A Bhounsule.
\newblock Heterogenous vehicle routing: comparing parameter tuning using
  genetic algorithm and bayesian optimization.
\newblock In {\em 2022 International Conference on Unmanned Aircraft Systems
  (ICUAS)}, pages 104--113. IEEE, 2022.

\bibitem{bahdanau2014neural}
Dzmitry Bahdanau, Kyunghyun Cho, and Yoshua Bengio.
\newblock Neural machine translation by jointly learning to align and
  translate.
\newblock {\em arXiv preprint arXiv:1409.0473}, 2014.

\bibitem{vaswani2017attention}
Ashish Vaswani, Noam Shazeer, Niki Parmar, Jakob Uszkoreit, Llion Jones,
  Aidan~N Gomez, {\L}ukasz Kaiser, and Illia Polosukhin.
\newblock Attention is all you need.
\newblock {\em Advances in neural information processing systems}, 30, 2017.

\bibitem{miller1960integer}
Clair~E Miller, Albert~W Tucker, and Richard~A Zemlin.
\newblock Integer programming formulation of traveling salesman problems.
\newblock {\em Journal of the ACM (JACM)}, 7(4):326--329, 1960.

\bibitem{mondal2023cooperative}
Md~Safwan Mondal, Subramanian Ramasamy, James~D. Humann, Jean-Paul~F.
  Reddinger, James~M. Dotterweich, Marshal~A. Childers, and Pranav~A.
  Bhounsule.
\newblock Cooperative multi-agent planning framework for fuel constrained
  uav-ugv routing problem, 2023.

\bibitem{ORtools}
Google.
\newblock {Google OR-tools}.
\newblock \url{https://developers.google.com/optimization}, 2021.
\newblock Online; accessed Feb 2, 2021.

\end{thebibliography}



\end{document}